\documentclass[conference]{IEEEtran}
\IEEEoverridecommandlockouts
\usepackage{cite}
\usepackage{amsmath,amssymb,amsfonts}
\usepackage{algorithmic}
\usepackage{graphicx}
\usepackage{textcomp}
\usepackage{tikz}
\usepackage{xcolor}
\usepackage{url}
\usepackage{balance}
\usepackage[capitalise, noabbrev]{cleveref}
\usepackage[nolist, nohyperlinks]{acronym}
\def\BibTeX{{\rm B\kern-.05em{\sc i\kern-.025em b}\kern-.08em
    T\kern-.1667em\lower.7ex\hbox{E}\kern-.125emX}}
\begin{document}

\title{
Federated Learning and Trajectory Compression for Enhanced AIS Coverage
\thanks{This work is partially funded by the EU's Horizon Europe research and innovation program under Grant No. 101070279 MobiSpaces.}
}

\author{\IEEEauthorblockN{Thomas Gr\"aupl, Andreas Reisenbauer,\\ Marcel Hecko, Anil Rasouli} \IEEEauthorblockA{\textit{Frequentis AG}, \\
Vienna, Austria \\
thomas.graeupl@frequentis.com \\
%ORCID: 0000-0002-7864-774X
}
\and
\IEEEauthorblockN{Anita Graser, Melitta Dragaschnig,\\ Axel Weissenfeld}
\IEEEauthorblockA{\textit{AIT Austrian Institute of Technology} \\
Vienna, Austria \\
anita.graser@ait.ac.at \\
%ORCID: 0000-0001-5361-2885
}
\and
\IEEEauthorblockN{Gilles Dejaegere, Mahmoud Sakr}
\IEEEauthorblockA{\textit{Université libre de Bruxelles} \\
Brussels, Belgium \\
gilles.dejaegere@ulb.be,\\ mahmoud.sakr@ulb.be}
}

\def\anita#1{{\sc \textcolor{blue}{Anita: }}{\sf \textcolor{red}{#1}}}
\def\mahmoud#1{{\sc \textcolor{blue}{Mahmoud: }}{\sf \textcolor{red}{#1}}}
\def\gilles#1{{\sc \textcolor{blue}{Gilles: }}{\sf \textcolor{red}{#1}}}
\def\thomas#1{{\sc \textcolor{blue}{Thomas: }}{\sf \textcolor{red}{#1}}}

\maketitle

\begingroup
\renewcommand\thefootnote{}% remove footnote marker
\footnotetext{%
© 2025 IEEE. Personal use of this material is permitted. Permission from IEEE must be obtained for all other uses, including reprinting/republishing this material for advertising or promotional purposes, creating new collective works for resale or redistribution to servers or lists, or reuse of any copyrighted component of this work in other works.
This is the \textbf{author’s accepted manuscript (AAM)} of the article:
“Federated Learning and Trajectory Compression for Enhanced AIS Coverage”.

The \textbf{Version of Record (VOR)} is published in the proceedings of the \textit{2025 Symposium on Maritime Informatics and Robotics (MARIS)} and is available at:
\url{https://doi.org/10.1109/MARIS64137.2025.11139533}

This accepted version is distributed under the terms of the Creative Commons Attribution 4.0 International License (CC BY 4.0).  
% in compliance with the open-access requirements of [Project Name, Grant No. XXXXX].%
}
\addtocounter{footnote}{-1}% reset counter
\endgroup

\begin{abstract}
This paper presents the VesselEdge system, which leverages federated learning and bandwidth-constrained trajectory compression to enhance maritime situational awareness by extending AIS coverage. VesselEdge transforms vessels into mobile sensors, enabling real-time anomaly detection and efficient data transmission over low-bandwidth connections. The system integrates the M³fed model for federated learning and the BWC-DR-A algorithm for trajectory compression, prioritizing anomalous data. Preliminary results demonstrate the effectiveness of VesselEdge in improving AIS coverage and situational awareness using historical data.
\end{abstract}


\section{Introduction}

% background
The \ac{AIS} is a tracking system that uses transceivers on ships to monitor marine traffic. Port authorities utilize AIS to assist vessels and track their movements. AIS is mandatory for ships with a gross tonnage of 300 or more and for all passenger ships, regardless of size.

% motivation
Ships equipped with AIS transceivers can be tracked by AIS base stations along coastlines. However, coastal AIS reception can be limited by the curvature of the earth, geographic conditions, or insufficient deployment. Additionally, as AIS usage has increased over time, some coastal areas experience high traffic density, which can result in the local overload of AIS radio channels \cite{IALA2017}.

% knowledge gap
It is therefore desirable to have a method to quickly extend AIS coverage to blind spots where needed, especially in critical situations such as \ac{SaR} operations. Cooperative SaR ships offer an opportunity for this, but the limited availability of high-speed communications at sea makes this technically challenging. It is impractical to relay \emph{all} AIS data from multiple vessels back to the control center over a low-bandwidth data connection. This necessitates reducing the amount of AIS data to be relayed to the coast. However, this must be done strategically, prioritizing relevant AIS messages i.e.~messages indicating movement anomalies.

% state-of-the-art
Anomaly detection from AIS is an active area of research~\cite{wolsing2022anomaly}. Graser et al. recently introduced M³fed~\cite{graser2024federated}, a \ac{FL} model for anomaly detection in the maritime domain. Other noteworthy FL contributions in different mobility domains include Koetsier's detection of anomalous vehicle trajectories at road intersections using federated learning~\cite{koetsier2022detection} and Li's multimodal transport demand forecasting network with attentive federated learning (MAFL)~\cite{li2023multimodal}.

% objectives
\vspace{1em}
In this paper, we propose an approach for extending AIS coverage using cooperative ships. The proposed approach involves a smart vessel-borne AIS pre-processing system called ``VesselEdge," which turns vessels into mobile sensors that extend the Vessel Traffic Services (VTS) center's range overcoming the limitations of coastal AIS' geographical coverage and the lack of high-speed communication networks at sea. The system combines machine learning based on M³fed with bandwidth-constrained trajectory compression methods.

\vspace{1em}
The remainder of this paper is structured as follows: First, we present the background of the proposed solution. Next, we describe the experimental setup and methods of VesselEdge, followed by the presentation of preliminary results. Finally, we provide an outlook on lab and sea trials scheduled for mid 2025.

\section{Background}

VesselEdge is part of the MobiSpaces project, funded by the EU's Horizon Europe research and innovation program. MobiSpaces aims to deliver a comprehensive, mobility-aware, and optimized data governance platform using \ac{ML} based mobility analytics to enhance the entire data path~\cite{Mobispaces2023}. VesselEdge is one of several use cases designed to validate the project. It demonstrates the \emph{mobility-aware learning at the edge} approach in a maritime setting by installing smart AIS receivers on moving vessels to collect data. This data is managed using the decentralized data management techniques developed in MobiSpaces. The validation aims to showcase improvements to VTS operations enabled by federated edge computing.

The goal of VesselEdge is to extend AIS coverage beyond the reach of coastal AIS antennas, especially in areas with limited or temporarily unavailable coverage, by deploying equipped vessels at sea. The vision is to enhance situational awareness for operators in control rooms, particularly for rescue mission coordinators who need visibility of the situation at the location of deployed first responder vessels with limited wireless connectivity, typically in the range of tens of kilobits per second~\cite{IALA2017}.

\begin{figure}[t]
	\centerline{\includegraphics[width=\linewidth]{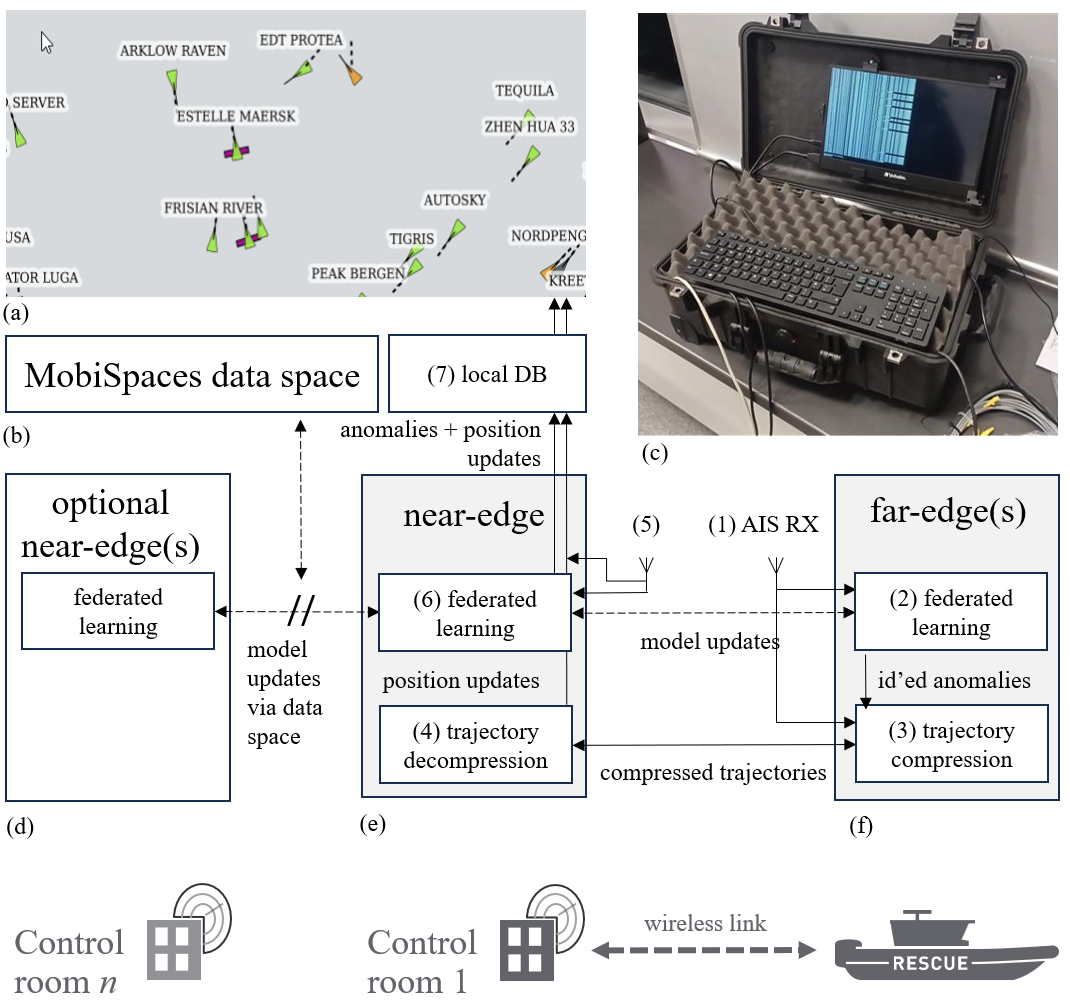}}
	\caption{VesselEdge aims to extend AIS coverage beyond coastal antennas by deploying equipped vessels at sea. The far-edge device (c and f) receives AIS messages and sends compressed, ML-prioritized trajectories to the near-edge infrastructure (a and e) at the coastal control room. Optionally, a second control room (d) can exchange ML updates with the first near-edge via the MobiSpaces data space (b).}
	\label{fig:setup}
\end{figure}

VesselEdge is intended for scenarios where a VesselEdge-equipped ship is deployed to a maritime emergency site, maintaining a low bit-rate wireless connection with a coastal control room as illustrated in \cref{fig:setup}. In this scenario, the proposed decentralized data management approach is as follows: (1) AIS data received by the deployed vessel is used to (2) continuously train a local machine learning model for movement anomaly detection on the edge. The ML model identifies unusual AIS data, and (3) a data management algorithm compresses the received AIS data into trajectories small enough for the low bit-rate wireless link while prioritizing anomalous data. The compressed AIS data is relayed to the coastal control center via the wireless link, where it is (4) decompressed and (7) injected into the local infrastructure for display on a tactical chart. Usually, the received data is merged with (5) the AIS data received on the coast. Optionally, (6) another instance of the ML model can be used to identify anomalies in all available AIS data on the coast to increase situational awareness. Finally, trained ML models can also be federated between control rooms and vessels of different organizations via the MobiSpaces data space~\cite{Mobispaces2023}. Federating ML models enables different organizational entities to share knowledge without sharing sensor data, which is often challenging.

\section{Experimental setup}

The VesselEdge experimental setup consists of two technology stacks illustrated in \cref{fig:setup}: (e) a near-edge server located in the coastal control center, and (c and f) a far-edge device installed on a mobile platform. These devices are interconnected via a maritime wireless data link. The end user, a coastal control officer, utilizes (a) the local infrastructure of the coastal control room to view a tactical chart.

Processes on both the near-edge and far-edge are deployed as containerized microservices, ensuring independence from specific devices or environments. In this setup, both near-edge and far-edge use the same hardware platform: HP Elite Mini 600 G9 i7 PCs with RHEL 7.9, hosting a Docker runtime.

\vspace{1em}
The far-edge device hosts the AIS receiver, specifically the \emph{easyRX21S-LAN-IS} AIS receiver. This receiver provides raw AIS data as a stream of NMEA sentences, which are decoded, cleaned, and converted to JSON format locally on the vessel. The JSON data stream is then consumed by the (2) (in \cref{fig:setup}) federated learning microservice to continuously train the local model for anomaly detection. The JSON messages and the results of the anomaly detection are ingested by the (3) trajectory compression microservice, which generates size-efficient trajectory representations prioritizing anomalous data. Multiple far-edge devices may be deployed.

Far-edge devices are connected to the near-edge via a maritime wireless link. For lab measurements, this is emulated for various communication channels, including low-bandwidth satellite links, VDES, or HF links with typical data rates \cite{IALA2017}. In sea trials, a commercial data link is used.

\vspace{1em}
The near-edge hosts two microservices: The (4) (in \cref{fig:setup}) trajectory decompression microservice reconstructs AIS messages from the received compressed data. Reconstructed data is stored in a local InfluxDB time-series database for logging and post-processing, and in a PostGIS database for visualizing the vessels on a tactical chart (a). The tactical chart is based on a commercial product. The (6) federated learning microservice runs a local instance of the ML model for anomaly detection, which may indicate anomalous vessel movements to the operator. The ML model used by the federated learning microservice may be the same model as used on the vessel or a larger model incorporating knowledge from supra-regional federated near-edges and control rooms. Federation is implemented with data exchange via the MobiSpaces data space~\cite{Mobispaces2023}.

\section{Method}
The methods for federated learning and trajectory compression are described in more detail below.

\subsection{Federated Learning} \label{sec:FL}

VesselEdge leverages \ac{FL} to train a machine learning model for detecting movement anomalies. FL is a machine learning approach where clients (e.g., mobile or edge devices) collaboratively train a model under the orchestration of a central server, while keeping the training data decentralized~\cite{Kairouz2021}. Specifically, VesselEdge uses the M³fed model~\cite{graser2024federated} for federated learning of movement anomaly detection models, aiming to improve data privacy and reduce communication costs. By applying M³fed, the amount of data transmitted from a FL client to the FL server can be reduced by approximately 98\% compared to centralized training (0.44GB versus 23.4GB for a year's worth of training with training updates once per day)~\cite{graser2024federated}. In the context of VesselEdge this applies to the exchange of models between control rooms via the MobiSpaces data space.

On the edge M³fed learns the expected distributions of vessel locations, speeds, and movement directions from historic AIS data and aggregates this information in so-called prototypes implemented as Gaussian Mixture Models. The model supports continuous learning from streaming data sources. The model needs to be pre-trained for the area of interest to ensure suitable anomaly detection performance. The training duration depends on the complexity of the movement patterns that are observed within the area. 

\subsection{Trajectory compression}

\begin{figure}[t]
	\centerline{\includegraphics[width=\linewidth]{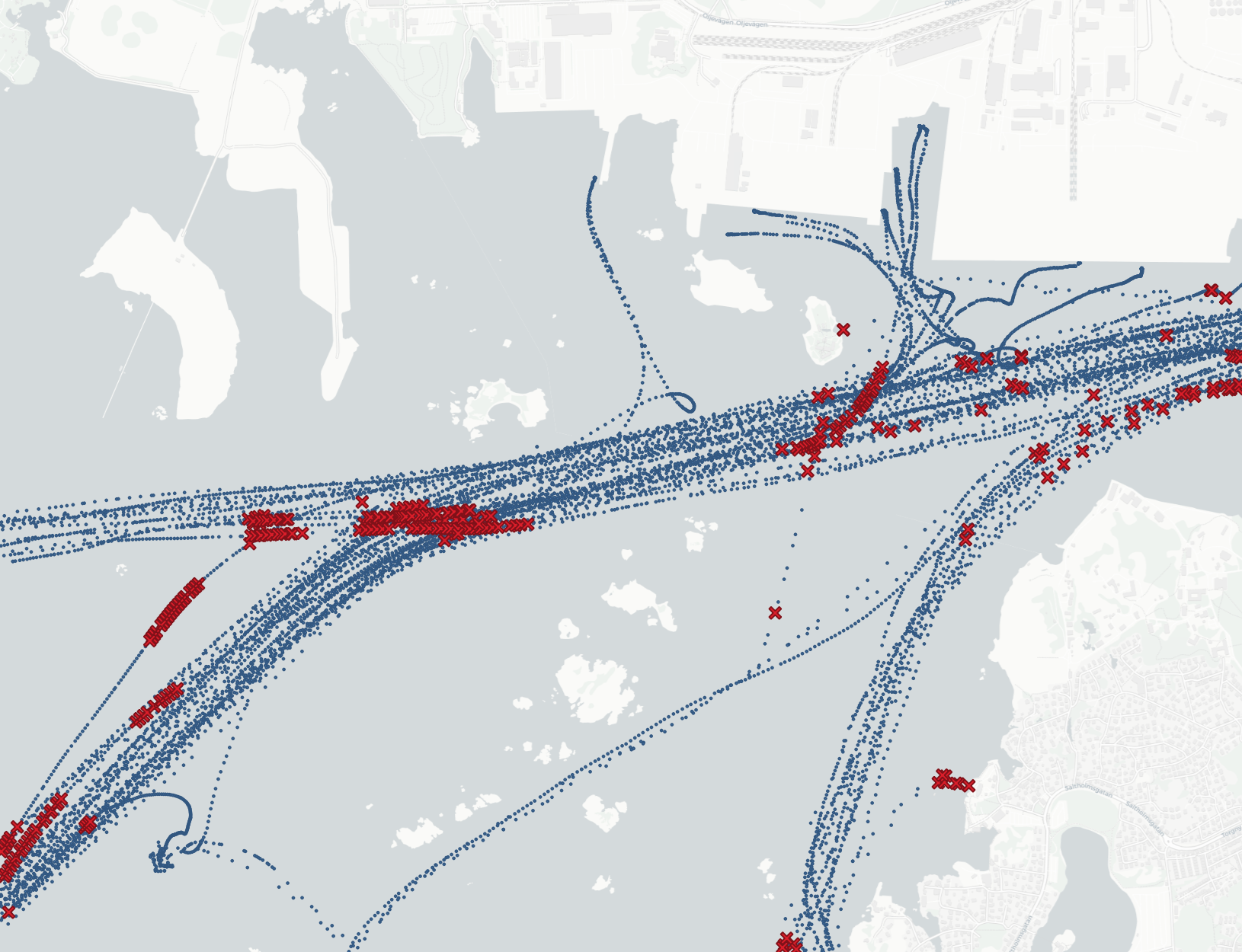}}
	\caption{AIS records (blue) and anomalies (red) detected by M³fed for tankers, cargo, and passenger vessels near Gothenburg harbor on 3rd July 2018.}
	\label{fig:anomalies}
\end{figure}

Existing trajectory compression algorithms try to balance the tradeoff between maximizing compression rate and minimizing data loss. In the context of this work, a novel challenge is the bandwidth constraint. As such the compression problem can be rephrased as follows: given a data size limitation, compress a set of trajectories to this data size while minimizing the data loss. To this end, the proposed bandwidth-constrained trajectory compression in VesselEdge modifies the classic \ac{DR} algorithm \cite{trajcevski2006line}. Originally, DR was designed to simplify trajectories in real-time, ensuring that the deviation between the actual position and its simplified trajectory never exceeds a given distance threshold. While this results in a smaller trajectory representation, the exact size of this representation cannot be predetermined.

VesselEdge addresses this limitation by imposing a bandwidth limit on the simplified trajectories instead of a deviation threshold. The modified algorithm, therefore, maintains a maximal number of vessel positions during a specified time-window. This modified algorithm is referred to as \ac{BWC-DR} \cite{dejaegere2025}. As in classical DR, BWC-DR is a simplification algorithm. This means that the simplified trajectory produced by this algorithm consists of a subset of the points of the initial trajectory, i.e., the subset of points that the BWC-DR assesses to be most relevant. The points of the initial trajectories are selected in the compressed representation by assigning them a priority proportional to the amount of distortion that would be introduced if that point was dropped from the compressed representation. Near-edge device can ``decompress" a trajectory by merging the compressed trajectories received from one or more far-edge devices.
 
\begin{figure}[t]
	\centerline{\includegraphics[width=\linewidth]{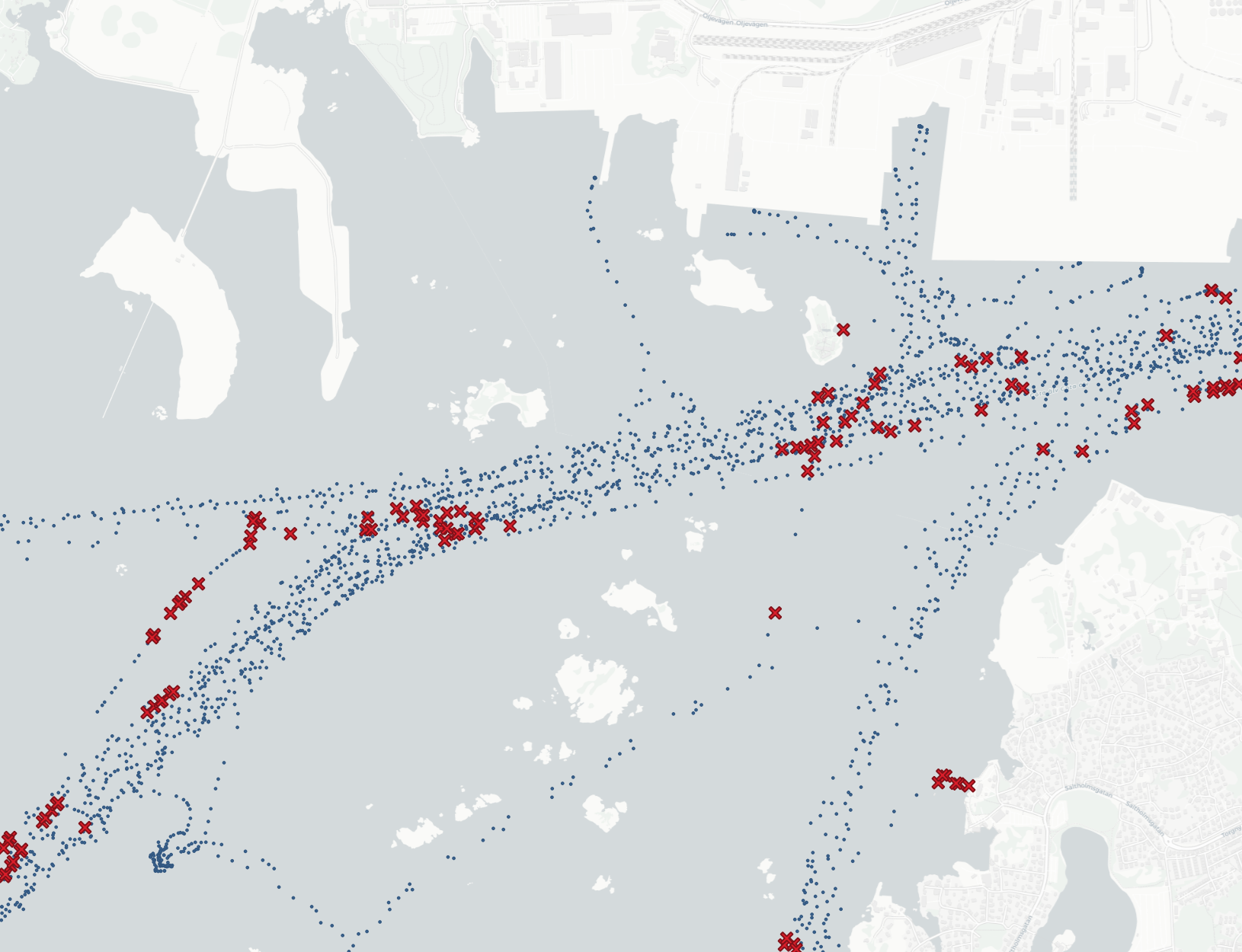}}
	\caption{AIS records (blue) and anomalies (red) post-compression with BWC-DR-A at bandwidth constraint that allows keeping $0.25$ of original points per window in average.}
	\label{fig:anomalies-compressed}
\end{figure}

For this work, the BWC-DR is further modified to exploit the results of FL. This modified version, \ac{BWC-DR-A}, increases the priority of the trajectory points of vessels with detected movement anomalies. This increase consists in multiplying the priorities of points of vessels for which a movement anomaly was detected in the three last time windows by a fixed factor (in this work 4). Clearly, this trades the overall compression of trajectories for the sake of enhancing the quality of the representation of vessels exhibiting abnormal behaviors, which are of greater interest to the application. \cref{fig:anomalies} and \cref{fig:anomalies-compressed} show the application of BWC-DR-A compression.

\section{Preliminary Results}

VesselEdge was preliminarily validated using historical AIS data\footnote{https://web.ais.dk/aisdata/} published by the Danish Maritime Authority. The data used to train the anomaly detection model is based on one year of AIS data collected between July 2017 and June 2018~\cite{graser2024federated}.
The results published in this paper were carried out with AIS data collected on July 3\textsuperscript{rd} 2018. 

\subsection{Anomaly Detection on the Far Edge}

In the first step, anomaly detection was performed on the far edge. This results in 5,920 anomalies across 72 distinct \acp{MMSI} out of 168,829 AIS records from 94 MMSIs. A subset of the identified anomalies is shown in Figure~\ref{fig:anomalies} to illustrate the results. While some anomalies appear as isolated records, most of them occur in sequences, referred to as \textit{anomaly events}.

\subsection{Trajectory compression}

\begin{figure}[t]
	\centerline{\includegraphics[width=\linewidth]{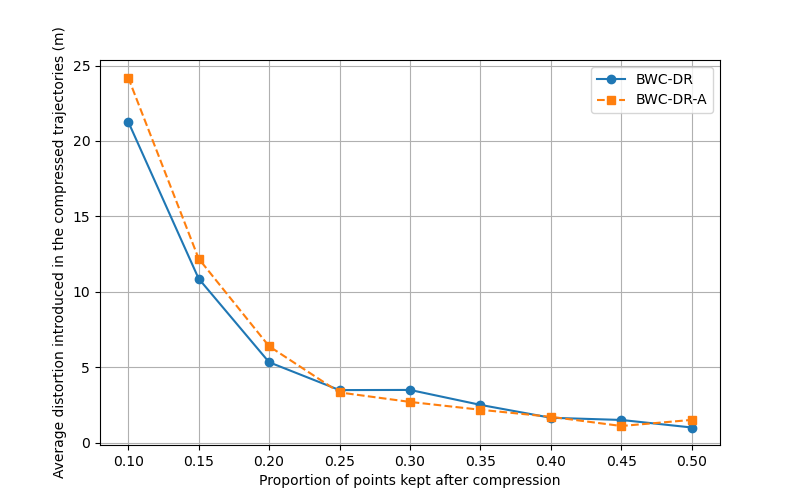}}
	\caption{Average distortion in meters depending on the bandwidth constraint introduced by the BWC-DR (blue) and anomaly-aware BWC-DR-A (orange) compression. The bandwidth constraint is expressed on the x-axis as the average fraction of points kept in a window after compression.}
	\label{fig:comp-dist}
\end{figure}

The compression component was evaluated based on the average distance between the simplified trajectories and the original ones. The compression time interval was set to 30 s. The bandwidth limitations was set from 10\% to 50\% of the AIS input available for the transmission. As it can be seen on Figure \ref{fig:comp-dist}, compressing the collected trajectories to 10\% of their original size, the BWC-DR algorithm introduces an average distortion of around 22 meters while BWC-DR-A introduces an average distortion of around 24 meters. This distortion drops to less than 5 meters for both methods if 25\% of the trajectory points are retained less than 2 meters if 50\% of the trajectory points are kept.

\subsection{Anomaly Detection on the Near Edge}

To evaluate the impact of trajectory compression on the detected anomalies, we compared the previously obtained anomaly detection results before compression with the anomalies obtained after compression. Table~\ref{tab1} shows how many of the 5,920 original anomalies are retained post-compression.

\begin{table}[htb]
    \caption{Anomalies after compression }
    \begin{center}
        \begin{tabular}{l|ccc}
            \textbf{Bandwidth constraint}  & \textbf{10\%}  & \textbf{25\%} & \textbf{50\%} \\
            \hline BWC-DR & 721 & 1616  &  2810  \\
            BWC-DR-A  &  1486  & 3045   & 4520
        \end{tabular}
        \label{tab1}
    \end{center}
\end{table}

Figure~\ref{fig:anomalies-compressed} displays the combined AIS trajectories and anomalies after compression. Compared to Figure~\ref{fig:anomalies}, it is evident that most locations previously marked as anomalies are successfully preserved post-compression.

\section {Conclusion \& Outlook}

This paper presented VesselEdge, a smart vessel-borne AIS pre-processing system designed to extend AIS coverage beyond the reach of coastal antennas. VesselEdge leverages decentralized machine learning on the edge and bandwidth-constrained trajectory compression methods to transform vessels into mobile sensors, enhancing situational awareness for maritime operations. The system was developed as part of the MobiSpaces project, which aims to deliver a mobility-aware and optimized data governance platform. 

Preliminary results indicate that the VesselEdge approach is promising, with successful evaluations of the individual components. However, additional work is required before the MobiSpaces project can be concluded. First, the federated learning and trajectory compression methods need to be evaluated at scale during lab trials under various bandwidth and AIS traffic settings. Second, an experimental setup involving a far-edge device will be deployed on a ship to demonstrate VesselEdge in a representative environment.

\begin{acronym}[parsep=0pt]
    \acro{AIS}{Automatic Identification System}
    \acro{CSS}{Coastal Surveillance System}
    \acro{DR}{Dead Reckoning}
    \acro{BWC-DR}{Bandwidth Constrained Dead Reckoning}
    \acro{BWC-DR-A}{Bandwidth Constrained Dead Reckoning with Anomalies}
    \acro{FL}{Federated Learning}
    \acro{GMDSS}{Global Maritime Distress and Safety System}
    \acro{ML}{Machine Learning}
    \acro{MMSI}{Maritime Mobile Service Identity}
    \acro{SaR}{Search and Rescue}
    \acro{VTS}{Vessel Traffic Services}
\end{acronym}

\vspace{1em}
\bibliographystyle{IEEEtran}
\bibliography{bibliography}

\end{document}